\pdfoutput=1
\documentclass{article}

\PassOptionsToPackage{numbers, compress}{natbib}
\usepackage[final]{neurips_2020}
\usepackage[utf8]{inputenc} 
\usepackage[T1]{fontenc}    
\usepackage{url}            
\usepackage{booktabs}       
\usepackage{amsfonts}       
\usepackage{amsmath}
\usepackage{nicefrac}       
\usepackage{microtype}      
\usepackage{graphicx}
\usepackage{soul,color}
\title{Learning Generalizable Physiological Representations from Large-scale Wearable Data}

\author{%
  Dimitris Spathis$^1$, Ignacio Perez-Pozuelo$^{2,3}$, Soren Brage$^2$,\\ \textbf{Nicholas J. Wareham}$^2$, and \textbf{Cecilia Mascolo}$^1$ \\
  
  \\
  $^1$Department of Computer Science and Technology, University of Cambridge, UK\\
  $^2$MRC Epidemiology Unit, School of Clinical Medicine, University of Cambridge, UK\\
  $^3$Alan Turing Institute, UK\\
   
}

\begin{document}

\maketitle

\begin{abstract} 
  To date, research on sensor-equipped mobile devices has primarily focused on the purely supervised task of human activity recognition (walking, running, etc), demonstrating limited success in inferring high-level health outcomes from low-level signals, such as acceleration. Here, we present a novel \textit{self-supervised} representation learning method using activity and heart rate (HR) signals without semantic labels. With a deep neural network, we set HR responses as the \textit{supervisory signal} for the activity data, leveraging their underlying physiological relationship. 
  
  We evaluate our model in the largest free-living combined-sensing dataset (comprising $>$280,000 hours of wrist accelerometer \& wearable ECG data) and show that the resulting embeddings can generalize in various downstream tasks through transfer learning with linear classifiers, capturing physiologically meaningful, personalized information. For instance, they can be used to predict (>70 AUC) variables associated with individuals’ health, fitness and demographic characteristics, outperforming unsupervised autoencoders and common bio-markers. Overall, we propose the first multimodal self-supervised method for behavioral and physiological data with implications for large-scale health and lifestyle monitoring. 

\end{abstract}

\section{Introduction}

The advent of wearable technologies has given individuals the opportunity to unobtrusively track everyday behavior. Given the rapid growth in adoption of internet-enabled wearable devices, sensor time-series comprise a considerable amount of user-generated data~\cite{blalock2016extract}. However, extracting meaning from this data can be challenging, since sensors measure low-level signals (e.g., acceleration) as opposed to the more high-level events that are usually of interest (e.g., arrhythmia, infection or obesity onset). 

 Deep learning has shown great promise in human activity recognition (HAR) tasks using wearable sensor data~\cite{yang2015deep,ma2019attnsense, alsheikh2015deep}, but relies on purely labeled datasets which are costly to collect ~\cite{bulling2014tutorial}. In addition, they are obtained in laboratory settings and hence might not generalize to free-living conditions where behaviours are more diverse, covering a long tail of activities~\cite{krishnan2018insights}. Multimodal learning has proven beneficial in supervised tasks such as fusing images with text to improve word embeddings~\cite{mao2016training}, video with audio for speech classification~\cite{ngiam2011multimodal}, or different sensor signals for HAR~\cite{radu2018multimodal}. However, all of these approaches rely on the modalities being used as parallel inputs, limiting the scope of the resulting representations. Self-supervised training allows for mappings of aligned coupled data streams (e.g. audio to images~\cite{owens2016ambient} or, in our case, activity to heart rate), using unlabeled data with supervised objectives~\cite{lan2019albert}. Physical activity is characterized by \textit{both} movement and the associated cardiac response to movement (e.g., heart rate increases after exercise and the dynamics of this increase are dictated by fitness levels~\cite{jones2000effect}), thus, leveraging these two signals concurrently likely produces better representations than either signal taken in isolation. Heart rate (HR) responses to exercise have been shown to be strongly predictive of cardiovascular disease (CVD), coronary heart disease (CHD) and all-cause mortality~\cite{savonen2006heart}.

\begin{figure*}
    \centering
    \includegraphics[width=1\linewidth]{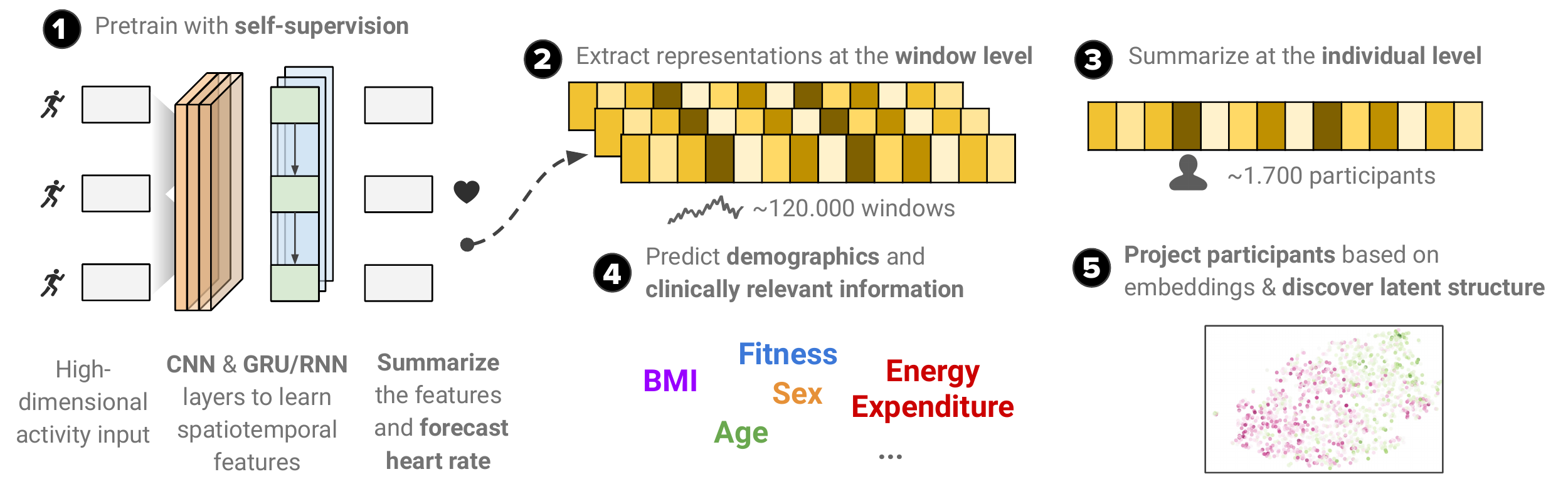}
	\caption{\textbf{Overview of model architecture and tasks}. 
	}
	\vspace{-0.6cm}
    \label{fig:model_hero}
\end{figure*}

 Here, we present \textit{Step2Heart}, a general-purpose self-supervised feature extractor for wearable data, which leverages the multimodal nature of modern wearable devices to generate participant-specific representations (see Fig. \ref{fig:model_hero}). This architecture can be broken into two parts: 1) First, a self-supervised pre-training task is used to forecast ECG-level quality HR in real-time by only utilizing activity signals, 2) then, we leverage the learned representations of this model to predict personalized health-related outcomes through transfer learning with \textit{linear} classifiers. We show that this mapping captures more meaningful information than autoencoders trained on activity data only or other bio-markers.

Traditionally, wearable sensors have captured activity counts, total number of steps or total volume of physical activity in minutes. Through our work we introduce a framework that can be used to extract physiologically meaningful, personalized representations from multimodal wearable sensors, moving a step beyond traditional intensity or step based measures. Previous work has explored representation learning using wearable data \cite{ballinger2018deepheart, ni2019modeling, saeed2019multi, sarkar2019self, hallgrimsson2018learning}, however, (with the exception of the work presented by \cite{hallgrimsson2018learning}), they have mostly focused on single sensor approaches modelled to infer one specific outcome. Our work is conducted using fine-grained (15'') multimodal signals and is used to infer a wide array of outcomes, ranging from demographic information to obesity or fitness indicators, towards facilitating the comprehensive monitoring of cardiovascular health and fitness at scale.

\section{Model}

\textbf{Problem formulation and notation.}  For this work, we assume $\mathcal{\textit{{N}}}$ samples, an input sequence $\mathcal{\textbf{X}}$ = (${\textbf{x}_{1}}$,...,${\textbf{x}_{N}}$) $\in \mathbb{R}^{N\times T\times F}$ and a target heart rate response  $\mathcal{\textbf{y}}$ = (${\textbf{y}_1}$,...,${\textbf{y}_{N}}$) $\in \mathbb{R}^{N}$. Additionally, we also consider contextual metadata like the hour of the day $\mathcal{\textbf{M}}$ = (${\textbf{m}_1}$,...,${\textbf{m}_{N}}$) $\in \mathbb{R}^{N\times F}$.
We use the same length $\mathcal{\textit{{T}}}$ for all sequences in our model. 
The intermediate representations of the model after training are $\mathcal{\textbf{E}}$ = (${\textbf{e}_1}$,...,${\textbf{e}_{N}}$) $\in \mathbb{R}^{N\times D}$ where $D$ is the latent dimension. These embeddings are aggregated at the user level $\mathcal{\tilde{\textbf{E}}}$ = ($\tilde{{\textbf{e}_1}}$,...,$\tilde{\textbf{e}_{N}}$) $\in \mathbb{R}^{\frac{N}{U} \times D}$, where $U$ is the number of users, in order to predict relevant outcome variables  $\mathcal{\tilde{\textbf{y}}}$ = ($\tilde{{\textbf{y}_1}}$,...,$\tilde{{\textbf{y}_{N}}}$) $\in \mathbb{R}^{N}$. We employ two representation learning tasks: self-supervised pre-training and a downstream transfer learning task. 

 {\bf Upstream task: self-supervised pre-training and HR forecasting.} Given the accelerometer input sensor sequence $\mathcal{\textbf{X}}$ and associated metadata $\mathcal{\textbf{M}}$, we predict the target HR $\mathcal{\textbf{y}}$ in the future. The input and target data shouldn't share temporal overlap in order to leverage the cardiovascular responses with the self-supervised paradigm by learning to predict the future \cite{wei2018learning}. Motivated by population differences in heart rates, we employ a custom \textit{quantile regression loss} (similar to \cite{rodrigues2018beyond, dabney2018distributional}) to account for the long-tails of the distribution (from sedentary to athletic people). 

 {\bf Downstream task: transfer learning of learned physiological representations.} Given the internal representations $\mathcal{\textbf{E}}$ --usually at the penultimate layer of the aforementioned neural network \cite{sanchez2019machine}--, we predict relevant variables $\mathcal{\tilde{{\textbf{y}}}}$ regarding the users' fitness and health using traditional classifiers (e.g. Logistic Regression). Inspired by the associations between word and document vectors in NLP \cite{le2014distributed}, we develop a simple aggregation method of sensor windows to the user level. The handling of sensor windows while taking the user into account is a common unsolved issue in the literature \cite{chen2019developing}.

\textbf{Model architecture}. The neural network builds upon architectures such as \textit{DeepSense}~\cite{yao2017deepsense}, which have been proven state of art in mobile sensing. It employs 2 CNN layers of 128 filters each, followed by 2 Bidirectional GRU stacked layers of 128 units each (resulting in 256 features due to bidirectional passes), and global element-wise average pooling. When using extra feature inputs, a \textit{ReLu} MLP of dimensionality 128 was employed for each one and its outputs were concatenated with the GRU output. Lastly, the final layer is a \textit{ReLu} MLP with a linear activation which is appropriate for continuous prediction tasks. We trained using the Adam \cite{kingma2014adam} optimizer for 300 epochs or until the validation loss stopped improving for 5 consecutive epochs. The quantiles we used for the pre-training loss were [0.01, 0.05, 0.5, 0.95, 0.99] so that they equally cover extreme and central tendencies of the heart rate distribution. For the Logistic Regression, the hyperparameters were found through 5-fold cross validation and were then applied to the test set.

\section{Data}

 \textbf{Study protocol.} Out of 12,435 men and women (aged 35-65) part of the \textit{Fenland Study}, a sub-sample of 2,100 were invited to a lab visit and was asked to wear a combined heart rate and movement chest sensor 
and a wrist accelerometer
on their non-dominant wrist for a week.
The study was approved by the University of Cambridge Ethics Committee. The \textit{chest ECG} measured heart rate and uniaxial acceleration in 15-second intervals 
while the \textit{wrist device} recorded 60 Hz triaxial acceleration. 
Participants were told to wear both monitors continuously 24/7. 
During a lab visit, all participants performed a treadmill test that was used to inform their $VO_{2}max$ (maximum rate of oxygen consumption and a gold-standard measure of fitness). Resting Heart Rate (RHR) was measured with the participant in a supine position using the \textit{chest ECG}.
These measurements were then used to calculate the Physical Activity Energy Expenditure (PAEE) \cite{brage2004branched}. 

\textbf{Pre-processing.} Accelerometer data underwent pre-processing consisting of: auto-calibration to local gravity, non-wear time detection and removal of participants with less than 72 hours of recording. 
Both the accelerometry and ECG signals were summarized to a common time resolution of one observation per 15 seconds and no further processing to the original signals was applied. Since time of the day can have a big impact on physical activity, we encoded the timestamps using \textit{cyclical temporal features} $T_f$ \cite{chakraborty2019advanced} by representing the month of the year and the hour of the day as $(x,y)$ coordinates on a circle. The intuition behind this encoding is that the model will ``see" that e.g. 23:59 and 00:01 are 2 minutes apart (not 24 hours). The signals were further segmented into \textit{non-overlapping} windows of 512 timesteps, each one comprising 15-seconds and therefore yielding a window size of approximately 2 hours. We divided the dataset into training and test sets randomly using an 80-20$\%$ split with the training set then being further split into training and validation sets (90-10$\%$), ensuring that the test and train set had disjoint user groups. Further, we normalized the data by performing min-max scaling on all features (sequence-wise for time-series and column-wise for tabular ones) on the training set and applied it to the test set. During training, the target data (HR bpm) is not scaled and the forecast is 15" into the future after the last activity input.

\section{Experimental results}

\textbf{Label and embeddings extraction.} For the transfer learning task, we studied if the embeddings $\mathbf{E}$ emerging from the self-supervised pre-training can predict outcomes ranging from demographics to fitness and health. After pre-processing, the analytical sample size encompassed 1506 participants with both sensor and lab visit data. To create binary labels we calculated the 50\% percentile in each variable's distribution on the training set and assigned equally sized positive-negative classes. Therefore, even continuous outcomes such as BMI or age become binary targets for simplification purposes (the prediction is high/low BMI etc). The window-level embeddings were averaged with an element-wise mean pooling to produce user-level embeddings. Then, to reduce overfitting, Principal Component Analysis (PCA) was performed on the training embeddings after standard scaling and the resulting projection was applied to the test set. We examined various cutoffs of explained variance for PCA, ranging from 90\% to 99.9\%. Intuitively, lower explained variance retained fewer components; in practice the number of components ranged from 10 to 160.

\textbf{Baselines.} A convolutional autoencoder was trained to compress the input data ($\mathbf{X}$ $\rightarrow$ $\mathbf{X}$) with a reconstruction loss. This unimodal baseline uses movement data only and is conceptually similar, albeit simpler, to \cite{aggarwal2019adversarial, saeed2019multi}. The intuition behind this is to assess whether \textit{Step2Heart} learns better representations due to learning a multimodal mapping of movement to heart rate ($\mathbf{X}$ $\rightarrow$ $\mathbf{y}$). To make a fair comparison, it has similar number of parameters to the self-supervised models and we use the bottleneck layer to extract embeddings (128 dimensions). Also, a univariate model trained with RHR only was tested.

\textbf{Effect of embeddings in generalization.} For this set of results, we use the best-performing model in forecasting HR, we extract embeddings, and train linear classifiers for different outcomes. Quantitatively, the embeddings achieved strong results in predicting variables like users' sex,  height, PAEE and weight (0.93, 0.82, 0.80 and 0.77 AUC respectively). Also, BMI, $VO_2max$ and age are moderately predictable (0.70 AUC). We also note the impact of the RHR in improving most results when added as input (A/R/T).

\begin{table}
\centering
\resizebox{0.90\textwidth}{!}{
\begin{tabular}{p{2cm}p{0.5cm}p{0.5cm}p{0.5cm}p{0.8cm}p{0.5cm}p{0.5cm}p{0.5cm}p{0.8cm}p{0.5cm}p{0.5cm}p{0.5cm}p{0.7cm}}
\toprule
Outcome & \multicolumn{12}{c}{AUC}                                                \\ \midrule
 & \multicolumn{4}{c}{Conv. Autoencoder}
        & \multicolumn{4}{|c|}{$\textit{Step2Heart}_{A/T}$}        & \multicolumn{4}{c}{$\textit{Step2Heart}_{A/R/T}$}  \\ \cmidrule(l){2-13} 
PCA\textbf{*}    & \multicolumn{1}{l}{$90\%$} & $95\%$ & $99\%$ & $99.9\%$ & $90\%$     & $95\%$    & $99\%$    & $99.9\%$ & $90\%$     & $95\%$    & $99\%$    & $99.9\%$      \\ \midrule

V$O_2 max$ & $52.6$  & $52.6$ & $59.6$ & $61.8$ & $58.6$ & $60$ & $63.9$ & $64.5$ & $\mathbf{68.3}$ & $67.8$& $68$& $68.2$     \\

PAEE & $69.6$  & $70.0$ & $70.2$ & $71.8$& $74.7$ & $74.7$ & $77.5$ & $76.8$ & $78.2$ & $79.2$& $\mathbf{80.6}$& $79.7$    \\

Height & $60.8$  & $60.3$ & $75.9$ & $79.4$& $66$ & $67.4$ & $77.4$ & $\mathbf{82.1}$ & $70.3$ & $74$& $80.5$& $81.3$   \\ 

Weight & $56.5$  & $56.2$ & $70.3$ & $72.1$& $65.7$ & $67.6$ & $75$ & $77.2$ & $69.9$ & $70.7$& $\mathbf{77.4}$& $76.9$   \\

Sex & $66.7$  & $67.0$ & $86.5$ & $89.7$& $72.3$ & $72.9$ & $87.1$ & $93.2$ & $76.2$ & $81.5$ & $91.1$ & $\mathbf{93.4}$   \\

Age & $46.2$  & $46.3$ & $53.9$ & $59.5$& $55.0$ & $61.7$ & $66.2$ & $66.9$ & $61.1$ & $63.8$ & $67.3$ & $\mathbf{67.6}$   \\

BMI & $51.6$  & $51.5$ & $60.1$ & $61.2$& $62.8$ & $63$ & $68.2$ & $67.6$ & $64.7$ & $66.1$ & $67.8$ & $\mathbf{69.4}$   \\

Resting HR & $49.1$  & $49.4$ & $55.8$ & $55.4$& $56.7$ & $56.6$ & $\mathbf{62.7}$ & $61.7$ & \multicolumn{4}{c}{N/A}   \\

\bottomrule  
\end{tabular}
}
\caption{\textbf{Transfer learning results}. Performance of embeddings in predicting variables related to health, fitness and demographic factors. A/R/T=Acceleration/RHR/Temporal features. (\textbf{*}percentage of explained variance by compressing the dimensionality of embeddings with PCA)}
\label{tab:transfer}
\vspace{-1.2cm}
\end{table}

\textbf{Impact of the new pre-training task.} Our results build on previous work ~\cite{hallgrimsson2018learning} however now using a multimodal and highly granular dataset. As a simple baseline, we followed their idea of using the RHR as a single predictor and we could not surpass an AUC of 0.55 for BMI and age. Also, the autoencoder baseline, which learns to compress the activity data, under-performs when compared to \textit{Step2Heart$_{A/T}$}, illustrating that the proposed task of mapping activity to HR captures physiological characteristics of the user, which translates to more generalizable embeddings. We note that both approaches operate only on activity data as inputs. This shows that the embeddings carry richer information than single bio-markers or modalities by leveraging the relationship between physical activity and heart rate responses. 

\textbf{Clinical relevance of results}. Obtaining these outcomes in large populations can be valuable for downstream health-related inferences which would normally be costly and burdensome (for example a $VO_2max$ test requires expensive laboratory treadmill equipment and respiration instruments). Additionally, PAEE has been strongly associated with lower risk of mortality in healthy older adults~\cite{manini2006daily,strain2020wearable}. Similarly, $VO_2max$ has been shown to be inversely associated with a higher risk of type-2 diabetes~\cite{katzmarzyk2005metabolic}.

\textbf{Impact of the latent dimensionality size.} From the representation learning perspective, we observe considerable gain in accuracy in some variables when retaining more dimensions (PCA components). More specifically, sex and height improve in absolute terms around +0.20 in AUC. However, this behavior is not evident in other variables such as PAEE and $VO_2max$, which seem robust to any dimensionality reduction. This implies that the demographic variables leverage a bigger dimensional spectrum of latent features than the fitness variables which can be predicted with a subsample of the features.  These findings could have great implications when deploying these models in mobile devices and deciding on model compression or distillation approaches~\cite{hinton2015distilling}.

\textbf{Discussion}. Through self-supervised learning, we can leverage unlabelled wearable data to learn meaningful representations that can generalize in situations where ground truth is inadequate or simply infeasible to collect due to high costs. Such scenarios are of great importance in mobile health where we may be able to achieve clinical-grade health inferences with widely-adopted devices. Our work makes contributions in the area of transfer learning and personalized representations, one of utmost importance in machine learning for health.

\begin{ack}
D.S was supported by the
Embiricos Trust Scholarship of Jesus College Cambridge, and EPSRC through Grant DTP (EP/N509620/1). I.P was supported by GlaxoSmithKline and EPSRC through an iCase fellowship (17100053). The authors declare that there is no conflict of interest regarding the publication of this work.
\end{ack}

\small
\bibliographystyle{plain}


\end{document}